\documentclass[a4paper,french]{rnti}

\usepackage[T1]{fontenc}
\usepackage[utf8]{inputenc}
\usepackage{amsmath}
\usepackage{subcaption}

\usepackage{url}

\usepackage{graphicx}
\usepackage{amsthm}

\titrecourt{Transformation en Shapelet Incertain - UST}

\nomcourt{M. F. MBOUOPDA et E. MEPHU NGUIFO}

\newtheorem{definition}{Définition}[subsection]

\DeclareMathOperator*{\argmax}{argmax}
\DeclareMathOperator{\UDISSIM}{UDISSIM}

\titre{Classification des Séries Temporelles Incertaines par Transformation Shapelet}

\auteur{Michael Franklin MBOUOPDA\affil{1},
        Engelbert MEPHU NGUIFO\affil{2}}

\affiliation{
    \affil{1}LIMOS, CNRS, University Clermont Auvergne, Clermont-Ferrand, France\\
          michael.mbouopda@uca.fr,\\
    \affil{2}LIMOS, CNRS, University Clermont Auvergne, Clermont-Ferrand, France\\
          engelbert.mephu\_nguifo@uca.fr\\
 }

\resume{%
La classification des séries temporelles est une tâche utilisée dans divers domaines tels que la météorologie, la médecine et la physique. Elle a pour but de classer des objets modélisés sous forme de séries temporelles. Des algorithmes toujours plus efficaces ont vu le jour durant cette dernière décennie pour accomplir cette tâche. Parmi ces algorithmes, il y a ceux basés sur la transformation en shapelet. Cependant aucun de ces derniers n'est applicable dans le cadre des séries temporelles incertaines. En utilisant les techniques de propagation de l'incertitude, nous proposons dans cet article une nouvelle mesure de dissimilarité incertaine que nous utilisons ensuite pour adapter la transformation en shapelet aux séries temporelles incertaines. Une validation expérimentale sur des données de la littérature montre l'intérêt de notre approche.
}

\summary{%
Time serie classification is used in a diverse range of domain such as meteorology, medicine and physics. It aims to classify chronological data. Many accurate approaches have been built during the last decade and shapelet transformation is one of them. However, none of these approaches does take data uncertainty into account. Using uncertainty propagation techiniques, we propose a new dissimilarity measure based on euclidean distance. We also show how to use this new measure to adapt shapelet transformation to uncertain time series classification. An experimental assessment of our contribution is done on some state of the art datasets.
}

\begin{document}

%
\section{Introduction}
Dans la dernière décennie il y a eu une grande explosion de la disponibilité de mesures dans une grande variété d'applications telles que la météorologie, l'astronomie et le traçage d'objets. Dans ces domaines, les mesures sont généralement représentées sous forme de séries temporelles \citep{uts_sim.1931D}, c'est à dire une séquence de données ordonnées sur une dimension temporelle. Également il y a eu un accroissement de techniques de classification des séries temporelles \citep{bagnall2017great,fawaz2019deep}. Cependant, les techniques existantes dans la littérature supposent toutes que les mesures sont faites de façon fiable. Nous savons pourtant qu'à toute mesure est associée une incertitude qui provient de sources diverses (instrument de mesure, environnement de mesure, etc). Cette incertitude n'est pas toujours négligeable, d'où la nécessité d'avoir des modèles qui prennent cela en considération.

La classification des séries temporelles par transformation Shapelet est parmi les techniques donnant les meilleures taux d'erreur en classification. Par ailleurs, elle est très appréciée par les experts du domaine car les résultats obtenus avec cette technique sont faciles à interpréter. Dans cet article, nous allons adapter cette technique pour la classification des séries temporelles incertaines. 

La classification par transformation en shapelet peut se résumer en trois étapes: premièrement il y a l'extraction des shapelets qui peut être vue comme une sélection des caractéristiques. Deuxièmement il y a la transformation shapelet qui consiste à calculer les vecteurs caractéristiques des différentes instances. Enfin il y a la classification proprement dite, faite avec un algorithme de classification supervisée quelconque. Dans les deux premières étapes, la distance euclidienne est utilisée pour mesurer la similarité entre séries temporelles; Or cette mesure de similarité ne prend par en compte l'incertitude. Notre contribution consiste premièrement à développer une mesure de similarité incertaine, et deuxièmement faire une classification par transformation Shapelet en utilisant cette mesure de similarité incertaine.

La suite de cet article est organisée comme suit: l'état de l'art est présenté à la section \ref{sec:related_works} et notre contribution est décrite à la section \ref{sec:contribution}. Dans la section \ref{sec:experiments} nous présentons les expérimentations et enfin, la section \ref{sec:conclusion} conclut ce papier et présente les perspectives.

\section{Etat de l'art}\label{sec:related_works}
La classification des séries temporelles a subit une très grande avancée durant cette dernière décennie. D'abord introduite avec les arbres de décision shapelet \citep{ye2009time}, elle a été généralisée en transformation shapelet \citep{hills2014STclassification}. Cette généralisation permet d'utiliser cette méthode de classification dans une modèle ensembliste appelé HIVE-COTE \citep{lines2018time}, qui est le meilleur modèle de classification de séries temporelles à l'heure actuelle. De loin que nous sachions, tous les modèles développés sont certes très performants, mais ont tous en commun le fait de ne pas prendre en compte l'incertitude dans les données.

La gestion de l'incertitude dans les séries temporelles s'est faite jusqu'ici par le développement de métriques capable de mesurer la similarité en prenant en compte l'incertitude. En effet plusieurs chercheurs s'y sont penchés et ont pu développer plusieurs mesures de similarité dites probabilistes \citep{uts_sim.1931D}. Ces mesures de similarité probabilistes calculent la probabilité que la similarité entre deux séries temporelles incertaines soient inférieure à un certain seuil défini par l'utilisateur. La mesure MUNICH \citep{munich2009} requiert plusieurs observations à chaque instant. Les mesures PROUD \citep{yeh2009proud} et DUST \citep{sarangi2010dust} quant à eux nécessitent que la distribution de l'incertitude soit connue. Contrairement à PROUD et DUST qui fournissent la probabilité que la similarité soit inférieure à un seuil, la mesure FOTS \citep{fots:siyou} calcule la dissimilarité. Par ailleurs FOTS n'a pas besoin de connaître la distribution de chaque observation, il n'a pas non plus besoin d'avoir plusieurs observations à un même instant comme MUNICH. Bien que FOTS résiste à l'incertitude, elle ne la prend pas en compte de façon explicite.

Les mesures de similarité probabilistes ne sont pas directement utilisables dans la transformation en Shapelet, car elles ne calculent pas la similarité proprement dite mais plutôt la probabilité que la similarité soit dans un intervalle. Par ailleurs, les informations qu'elles requièrent ne sont pas toujours disponibles; En effet la distribution de l'incertitude n'est pas toujours connue (PROUD et DUST) et on n'a pas toujours plusieurs observations à chaque instant (MUNICH). FOTS pourrait être une solution, mais en plus de ne pas prendre en compte l'incertitude de façon explicite, sa complexité quadratique rendrait la transformation Shapelet encore plus lente qu'elle ne l'est déjà. Dans ce papier nous proposons une métrique qui ne requiert aucune information supplémentaire et dont la complexité est linéaire.

\section{Contribution}\label{sec:contribution}
\subsection{Motivation}
Contrairement à l'erreur qui peut être supprimée en étant minutieux, l'incertitude ne peut pas être supprimée \citep{err_analysis_jrtaylor}. Quelque soit les techniques utilisées pour faire une mesure, il y a toujours de l'incertitude. Ainsi la similarité entre deux mesures incertaines ne peut être données de façon fiable. De même, la similarité entre deux séries temporelles incertaines doit être accompagnée d'une incertitude. Bien qu'il existe plusieurs façons de représenter l'incertitude, nous allons nous limiter à la représentation donnée par \citet*{err_analysis_jrtaylor}.

\begin{definition}
	Une mesure incertaine $x$ est définie par une estimation optimiste $x_{opt}$ et une incertitude $\delta{x}$ sur cette estimation. On note:
	\[
		x = x_{opt} \pm \delta{x}
	\]
\end{definition}

Cette définition signifie que la valeur fiable (si l'incertitude était éliminée) de la mesure se trouve dans l'intervalle $[x_{opt} - \delta{x}, x_{opt} + \delta{x}]$.

La distance euclidienne est une métrique simple utilisée pour mesurer la similarité entre les séries temporelles lorsqu'il y a pas d'incertitude. Elle est particulièrement très utilisée dans les méthodes à bases de shapelets \citep{ye2009time,hills2014STclassification,bagnall2017great}. Notre contribution consiste à utiliser les techniques de propagation de l'incertitude \citep{err_analysis_jrtaylor} pour définir une nouvelle mesure de similarité incertaine, basée sur la distance euclidienne. 

\subsection{Mesure de Dissimilarité Incertaine}
Commençons par rappeler la définition du carré de la distance euclidienne :
\begin{definition}
	Soit deux vecteurs $V=(v_1, v_2, ..., v_n)$ et $U=(u_1, u_2, ..., u_n)$. Le carré de la distance euclidienne (\textit{ED}) entre ces deux vecteurs est définie comme suit:
	\[
		ED(V,U) = \sum_{i=1}^{n}(v_i-u_i)^2
	\]
\end{definition}

Nous souhaitons adapter cette définition pour des vecteurs incertains; un vecteur incertain étant un vecteur dont les composantes sont des mesures incertaines. Un vecteur incertain $V$ est noté comme suit:
\[
	V \pm \delta{V} = (v_1 \pm \delta{v_1}, v_2 \pm \delta{v_2}, ..., v_n \pm \delta{v_n})
\]

Les opérateurs utilisés dans la distance euclidienne sont les opérateurs d'addition, de soustraction et de puissance. Lorsque ces opérateurs sont utilisés sur des mesures incertaines, le résultat obtenu a une incertitude que l'on peut quantifier. En effet, soient $x=x_{opt} \pm \delta{x}$ et $y=y_{opt} \pm \delta{y}$ deux mesures incertaines, \citet{err_analysis_jrtaylor} nous donne les propriétés suivantes:
\begin{itemize}
	\item L'addition de $x$ et $y$ donne $z=z_{opt} \pm \delta{z}$, avec $z_{opt}=x_{opt}+y_{opt}$ et $\delta{z}=\delta{x}+\delta{y}$
	\item La soustraction de $x$ et $y$ donne $z=z_{opt} \pm \delta{z}$, avec $z_{opt}=x_{opt}-y_{opt}$ et $\delta{z}=\delta{x}+\delta{y}$
	\item La puissance $n$-ième de $x$ donne $z=z_{opt} \pm \delta{z}$, avec $z_{opt}=(x_{opt})^n$ et $\delta{z}=|n| \frac{\delta{x}}{|x|} |(x_{opt})^n|$
\end{itemize}

En utilisant ces propriétés, nous avons pu propager l'incertitude dans le calcul du carré de la distance euclidienne entre deux vecteurs incertains. Le résultat obtenu est une mesure de dissimilarité, que nous avons appelé \textit{UDISSIM}

\begin{definition} 
	La dissimilarité entre deux vecteurs incertains $V \pm \delta{V} = (v_1 \pm \delta{v_1}, v_2 \pm \delta{v_2}, ..., v_n \pm \delta{v_n})$ et $U \pm \delta{U} = (u_1 \pm \delta{u_1}, u_2 \pm \delta{u_2}, ..., u_n \pm \delta{u_n})$ est définie comme suit:
	\[
		\UDISSIM(V \pm \delta{V}, U \pm \delta{U}) = \sum_{i=1}^{n}(v_i-u_i)^2 \pm 2\sum_{i=1}^{n}|v_i-u_i| \times (\delta{v_i} + \delta{u_i})
	\]
\end{definition}

On peut remarquer que l'estimation optimiste de \textit{UDISSIM} vaut exactement la distance euclidienne; ainsi si l'incertitude est nulle alors, \textit{UDISSIM} est équivalent à \textit{ED}.

Afin d'utiliser \textit{UDISSIM} comme mesure de dissimilarité dans les shapelets, il est nécessaire de pouvoir comparer deux valeurs incertaines. Lorsque une même mesure est réalisée par plusieurs personnes ou par plusieurs instruments, la notion de discordance (<< discrepancy >> en anglais) permet de savoir si les différentes mesures sont en accord ou pas \citep{err_analysis_jrtaylor}. Cette notion n'est pas suffisante pour mesurer la similarité, car on ne cherche pas juste un élément similaire, mais généralement l'élément le plus similaire. De ce fait, il faut une relation d'ordre pour l'ensemble des mesures incertaines. Cette relation d'ordre est données avec les deux définitions suivantes:

\begin{definition}[Égalité]
	Deux mesures incertaines sont égales si et seulement si leurs estimations optimistes et leurs incertitudes sont égales.
	\[
		x = y \iff x_{opt} = y_{opt} \land \delta{x} = \delta{y}
	\]
\end{definition}

\begin{definition}[Infériorité]\label{prop:inf}
	La plus petite de deux mesures incertaines est celle dont l'estimation optimiste est la plus petite. Si leurs estimations optimistes sont égales, alors c'est celle avec la plus petite incertitude qui est la plus petite.
	\[
		x<y \iff(x_{opt} < y_{opt}) \lor ((x_{opt} = y_{opt}) \land (\delta{x} < \delta{y}))
	\]
\end{definition}

La propriété d'infériorité s'explique par le fait que nous accordons premièrement une certaine confiance à l'estimation optimiste. Deuxièmement nous avons une plus grande confiance aux mesures dont l'incertitude est plus petite. La relation d'ordre telle que nous l'avons définie ne donne pas toujours le bon résultat. En effet, si $ x = 2 \pm 0.5 $ et $ y = 2 \pm 0.1 $; d'après la définition de l'infériorité, $ y < x $. S'il existait une technique permettant de faire une mesure exacte de $x$ et de $y$, on pourrait très bien avoir $x=1.5$ et $y=2.1$. L'ordre donné par notre définition serait donc incorrect.

Afin d'évaluer expérimentalement la pertinence de \textit{UDISSIM}, nous l'avons utilisée en lieu et place de \textit{ED} pour faire la classification de séries temporelles incertaines en utilisant la transformation shapelet. Dans la section suivante, nous présenterons les expérimentations que nous avons conduites, ainsi que les résultats obtenus.

\subsection{Transformation en shapelet incertain: UST}
L'algorithme de classification par transformation shapelet est décrit par \citet{hills2014STclassification}. Cet algorithme est basé sur le calcul de la similarité entre un Shapelet et une séries temporelles. Dans cette section, nous allons présenter <<Uncertain Shapelet Transform (UST)>>, une adaptation de l'algorithme de transformation shapelet pour les séries temporelles incertaines. Pour ce faire, il est nécessaire de définir la notion de Shapelet incertain.

\begin{definition}[Série temporelle incertaine]
	Une série temporelle $T$ est une suite chronologique de $m$ mesures incertaines. $m$ est appelée la taille de la série.
	\[
		T = T_{opt} \pm \delta{T} = \{t_1 \pm \delta{t_1}, t_2 \pm \delta{t_2}, ..., t_m \pm \delta{t_m}\}
	\]
\end{definition}

\begin{definition}[Sous-séquence incertaine]
	Une sous-séquence incertaine $S$ d'une série temporelle incertaine $T$  est une suite ordonnée de $l$ mesures incertaines consécutives de $T$.
	\[
		S = S_{opt} \pm \delta{S}  = \{t_{i+1} \pm \delta{t_{i+1}}, ..., t_{i+l} \pm \delta{t_{i+l}}\}
	\]
\end{definition}

\begin{definition}[Dissimilarité entre deux sous-séquences incertaines]
	La dissimilarité entre deux sous-séquences incertaines $S$ et $R$ de même taille est donnée par:
	\[
		d = \UDISSIM(S,R) = \UDISSIM(R,S).
	\]
\end{definition}

\begin{definition}[Dissimilarité entre sous-séquence et série temporelle incertaines]
	La dissimilarité entre une sous-séquence incertaine $S$ et une série temporelle incertaine $T$ est la dissimilarité incertaine entre $S$ et la sous-séquence de $T$ de même taille que $S$ qui soit la plus proche de $S$.
	\[
		d = \UDISSIM(S,T)=\UDISSIM(T,S)=\min \{ \UDISSIM(S,R) \ \forall R \in T\}
	\]
\end{definition}

\begin{definition}[Séparateur incertain]
	Un séparateur incertain $sp$ d'un ensemble de séries temporelles $D$ est une sous-séquence incertaine qui le divise en deux sous ensembles $D_1$ et $D_2$, avec:
	\[
		D_1 = \{ T\: | \UDISSIM(T,sp) \le \epsilon\:, \forall T \in D \}, D_2=\{ T\: | \UDISSIM(T,sp) > \epsilon\:, \forall T \in D \}
	\]
\end{definition}

Tout comme dans la transformation shapelet classique (sans prise en compte de l'incertitude), la qualité d'un séparateur est donnée par le gain d'information(IG) \citep{hills2014STclassification}, et se calcul sous la base de l'entropie.

En utilisant les définitions précédentes, nous pouvons  désormais définir formellement ce qu'est un Shapelet incertain. 
\begin{definition}[Shapelet incertain]
	Un shapelet incertain $S$ pour un jeu de données $D$ de séries temporelles incertaines est un séparateur incertain de $D$ qui maximise le gain d'information.
	\[
		S=\argmax_{sp}(IG(D,sp))
	\]
\end{definition}

Étant donné un jeux données de séries temporelles incertaines, la classification de séries temporelles incertaines par transformation shapelet se fait en trois étapes:
\begin{enumerate}
	\item Extraction des $k$ meilleurs shapelets incertains $S$. Ici, \textit{UDISSIM} est utilisée pour identifier les meilleurs séparateurs du jeu de données.   
	\item Calcul des vecteurs caractéristiques: Pour chaque série temporelle incertaine $T_i$ dans le jeu de données, son vecteur caractéristique $V_i$ est un vecteur de taille $k$ dont la composante $j$ est $V_j=\UDISSIM(T_i, S_j)$
	\item Classification proprement dite: Un modèle de classification supervisé est entraîné sur l'ensemble des couples $(V_i, c_i)$, où $V_i$ et $c_i$ sont respectivement le vecteur caractéristique et la classe de la série temporelle incertaine $T_i$. Ce modèle doit faire la classification en prenant en compte l'incertitude qui est présente dans les vecteurs caractéristiques.
\end{enumerate}
Il s'agit là de l'algorithme de classification par transformation shapelet \citep{hills2014STclassification} dans lequel l'incertitude est prise en compte. En effet, nous utilisons dans la phase d'extraction des shapelets \textit{UDISSIM} en lieu et place de \textit{ED}, ce qui nous permet de propager l'incertitude pendant la transformation (calcul des vecteurs caractéristiques). Enfin, le modèle de classification est entraîné en étant conscient de l'incertitude dans les données. Dans la partie suivante, nous présentons les expérimentations et les résultats que nous obtenons avec notre approche. Nous avons appelé notre algorithme <<Uncertain Shapelet Transform>> ou <<UST>>

\section{Expérimentations}\label{sec:experiments}
\subsection{Jeux de données}
Afin d'évaluer notre contribution, nous avons fait des expérimentations en utilisant les $21$ jeux de données listés dans le tableau \ref{tab:datasets}. Ils proviennent du célèbre dépôt UEA/UCR{\footnote{\url{http://www.timeseriesclassification.com/dataset.php}}}. Étant donné que ces jeux de données sont supposés fiables (les mesures sont sans incertitudes), nous y avons ajouté de l'incertitude suivant une distribution normale. Pour chaque jeu de données, la distribution normale de l'incertitude a une moyenne nulle et un écart-type égal à $ \sigma $, $ \sigma $ étant l'écart-type sur le jeu de données. L'ensemble d'apprentissage utilisé par notre modèle est l'ensemble des jeux de données incertains obtenues et la valeur absolu des incertitudes générées.

\begin{table}[!h]
	\centering
	\begin{tabular}{|c|c|c|c|c|}
		\hline
		\textbf{Dataset} & \textbf{Nb d'instances train/test} & \textbf{longueur des séries} & \textbf{Nb classes}\\
		\hline
		Chinatown & 20/345 & 25 & 2\\
		SmoothSubspace & 150/150 & 16 & 3\\
		ECGFiveDays & 23/861 & 137 & 2\\
		SonyAIBORobotSurface & 120/601 & 71 & 2\\
		MoteStrain & 20/1252 & 85 & 2\\
		Fungi & 18/186 & 202 & 18\\
		DiatomSizeReduction & 16/306 & 346 & 4\\
		ArrowHead & 36/175 & 252 & 3 \\
		TwoLeadECG & 23/1139 & 83 & 2 \\ 
		BirdChicken & 20/20 & 513 & 2\\
		PowerCons & 180/180 & 145 & 2\\
		UMD & 36/144 & 151 & 3\\
		CBF & 30/900 & 129 & 3\\
		BME & 30/150 & 129 & 3\\
		GunPoint & 50/150 & 151 & 2 \\
		MoteStrain & 20/1252 & 85 & 2\\ 
		ArrowHead & 36/175 & 252 & 3 \\
		Plane & 105/105 & 145 & 7 \\ 
		DistalPhalanxTW & 400/139 & 81 & 6 \\         
		GunPointOldVersusYoung & 136/315 & 151 & 2\\
		SyntheticControl & 300/300 & 61 & 6 \\
		DistalPhalanxTW & 400/139 & 81 & 6 \\
		MelbournePedestrian & 1200/2450 & 25 & 10\\
		MiddlePhalanxOutlineAgeGroup & 400/154 & 81 & 3\\
		MiddlePhalanxOutlineCorrect & 600/291 & 81 & 2\\
		MiddlePhalanxTW & 399/154 & 81 & 6\\
		PowerCons & 180/180 & 145 & 2\\
		ProximalPhalanxOutlineAgeGroup & 400/205 & 81 & 3\\
		ProximalPhalanxTW & 400/205 & 81 & 6\\
		\hline
	\end{tabular}
	\caption{Liste des jeux de données}
	\label{tab:datasets}
\end{table}

\subsection{Code Source}
Nous avons réutilisé le code open source mise à disposition par \citet{bagnall2017great} à cette adresse\footnote{ \url{https://github.com/uea-machine-learning/tsml}}. Ce dépôt github contient une implémentation en langage Java de plusieurs algorithmes de classification des séries temporelles parmi lesquels l'algorithme de classification par transformation shapelet. 

Nous avons étendu le code en y implémentant UDISSIM et UST. Nous avons essayé autant qu'on a pu de réutiliser le code existant car il est très bien fait. Nous avons mis le code source étendu à la disposition du grand public à l'adresse suivante: \url{https://github.com/frankl1/Uncertain-Shapelet-Transform}.

En ce qui concerne l'ajout de l'incertitude dans les données, nous avons utilisé un notebook Jupyter. Le notebook est écrit en langage Python (Version 3) et est disponible à l'adresse suivante: \url{https://github.com/frankl1/Uncertain-Shapelet-Transform/blob/master/add-noise.ipynb}

\subsection{Résultats}
Nous avons comparé deux modèles UST. Le premier (UST-DT) utiliser un arbre de décision pour la classification des vecteurs caractéristiques, tandis que le second (UST-ROTF) utilise plutôt les forêts de rotation (<<Rotation Forest>> en anglais). Les vecteurs caractéristiques obtenus avec la propagation ayant de l'incertitude, le modèle de classification devrait donc être capable de prendre cela en compte. Cependant, cette problématique n'est pas le but de cet article. Pour nos expérimentations, nous avons défini deux façons basiques nous permettant de prendre en compte l'incertitude.

\begin{figure}
	\centering
	\begin{subfigure}{.5\textwidth}
		\centering
		\includegraphics[width=.7\linewidth]{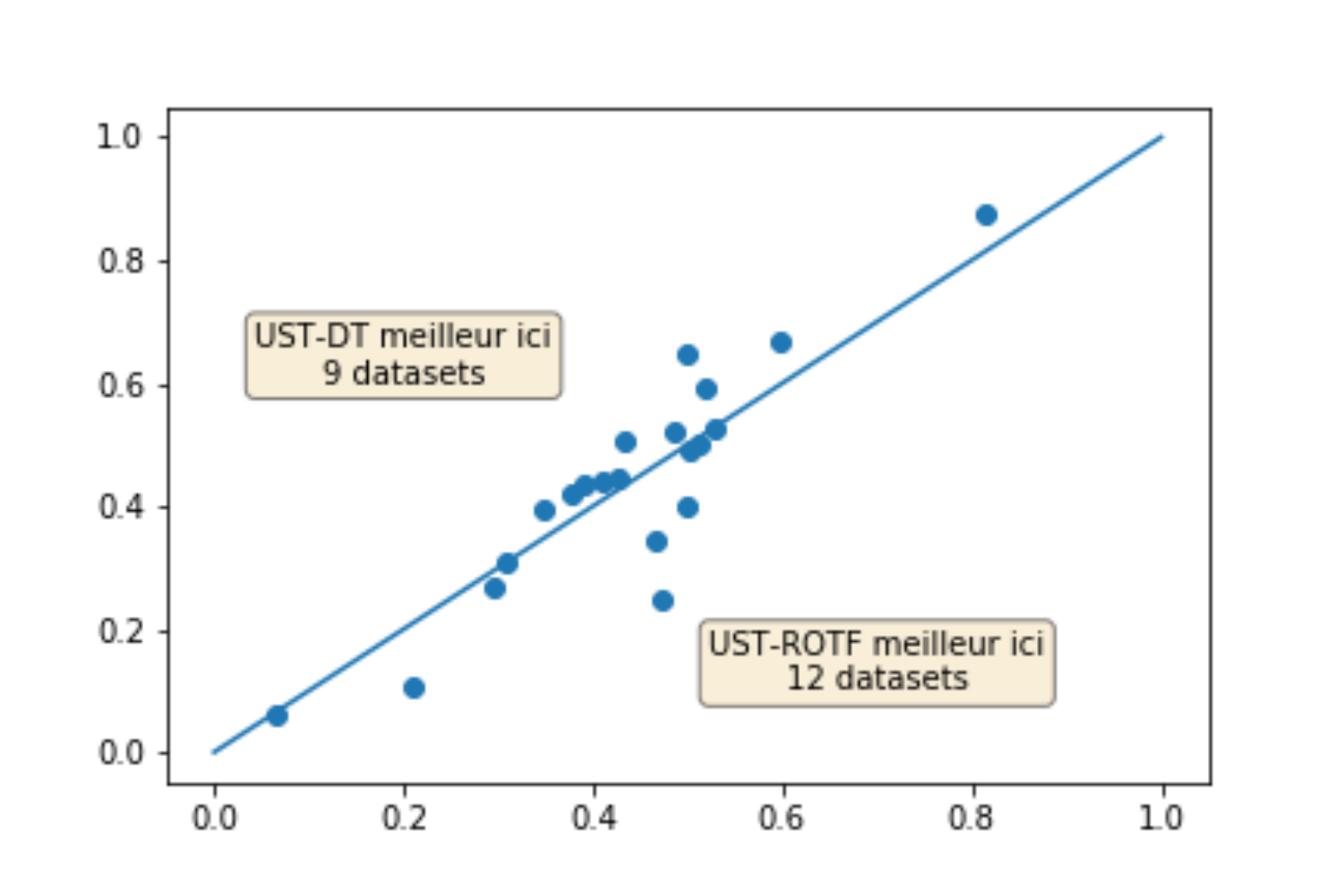}
		\caption{UST-DT vs UST-ROTF}
		\label{fig:scatter_plot_ust_dt_vs_rotf_flat}
	\end{subfigure}%
	\begin{subfigure}{.5\textwidth}
		\centering 
		\includegraphics[width=.7\linewidth]{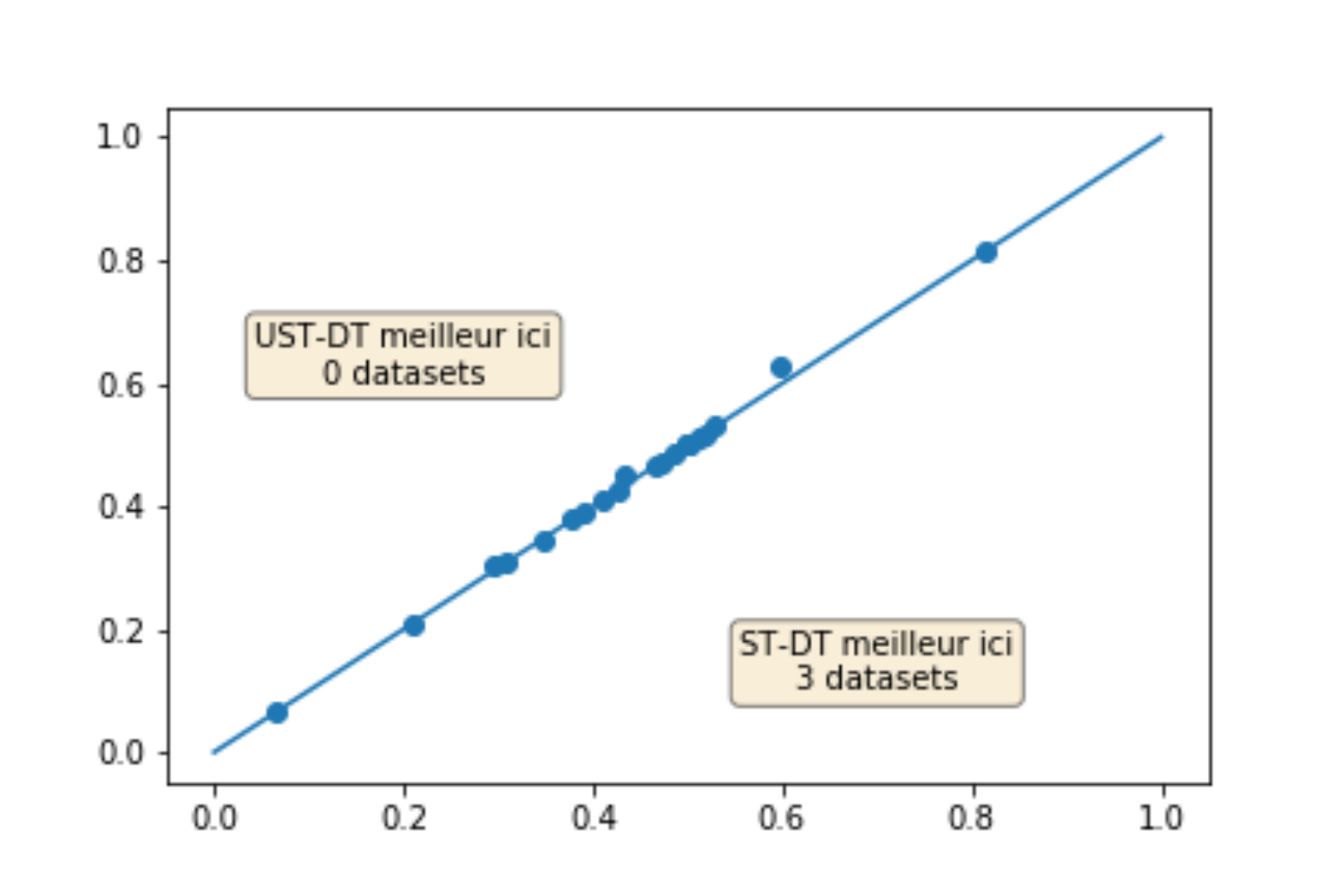}
		\caption{UST-DT vs ST-DT}
		\label{fig:scatter_plot_ust_vs_st_dt_flat}
	\end{subfigure}
	\caption{Taux de classification avec vecteurs caractéristiques aplatis}
\end{figure}

Dans la première, les vecteurs caractéristiques sont <<aplatis>>; c'est-à-dire que la première moitié contient les estimations optimistes et la seconde moitié les incertitudes. Ainsi l'incertitude est vue par le modèle de classification comme une caractéristique à part entière et non comme une méta-caractéristique. La figure \ref{fig:scatter_plot_ust_dt_vs_rotf_flat} présente un nuage de points représentant les précisions obtenues par les deux modèles sur les différents jeux de données. UST-DT est meilleur sur $9$ jeux de données, UST-ROTF sur $12$. 

La figure \ref{fig:scatter_plot_ust_vs_st_dt_flat} compare UST à ST(transformation shapelet classique sans prise en compte de l'incertitude). Les deux modèles donnent les mêmes précisions sur $18$ jeux de données. Sur les $3$ restants, le modèle classique donne une précision supérieure de $0.8\%$ en moyenne.

\begin{figure}
	\centering
	\begin{subfigure}{.5\textwidth}
		\centering
		\includegraphics[width=.7\linewidth]{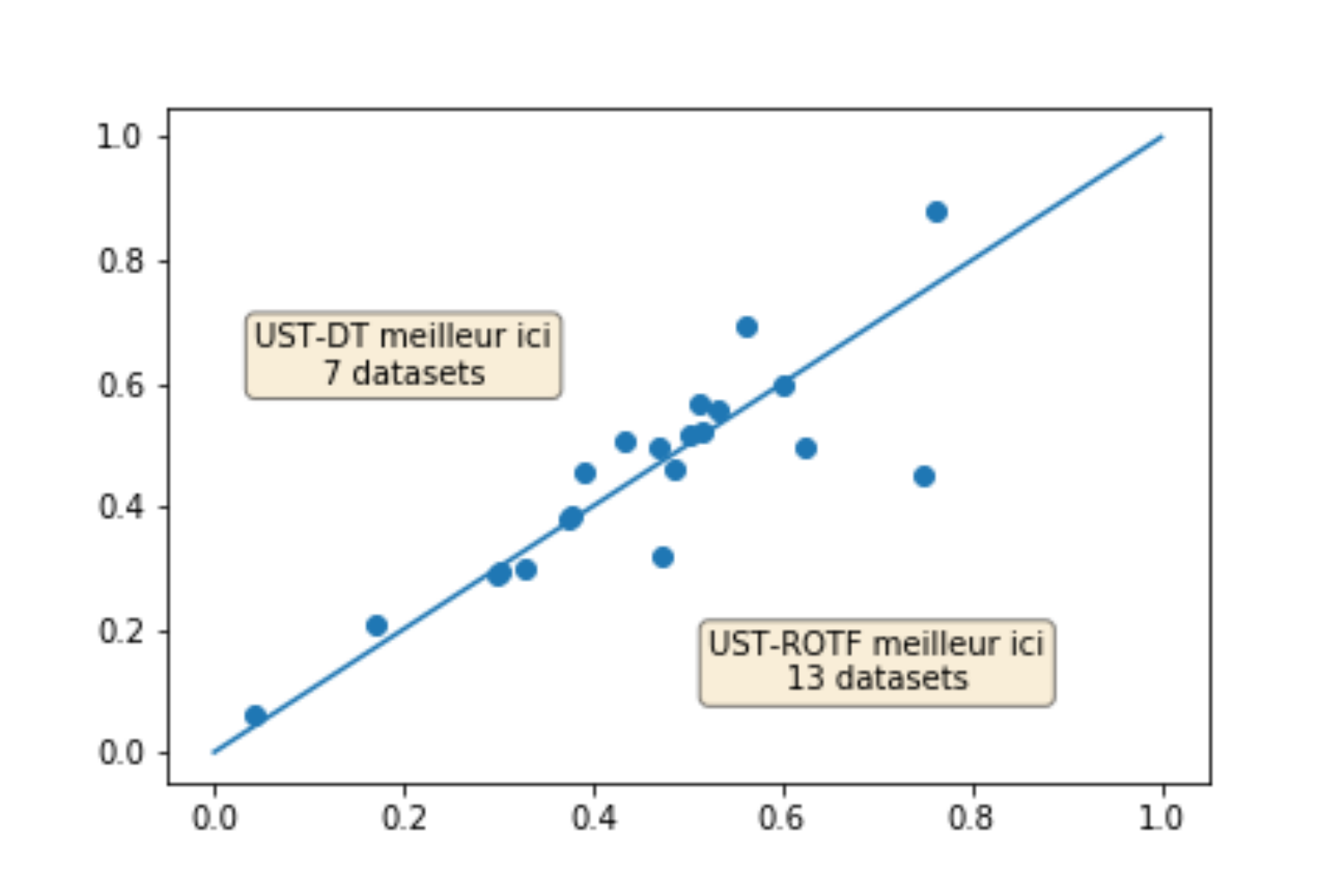}
		\caption{UST-DT vs UST-ROTF}
		\label{fig:scatter_plot_ust_dt_vs_rotf_gauss}
	\end{subfigure}%
	\begin{subfigure}{.5\textwidth}
		\centering 
		\includegraphics[width=.7\linewidth]{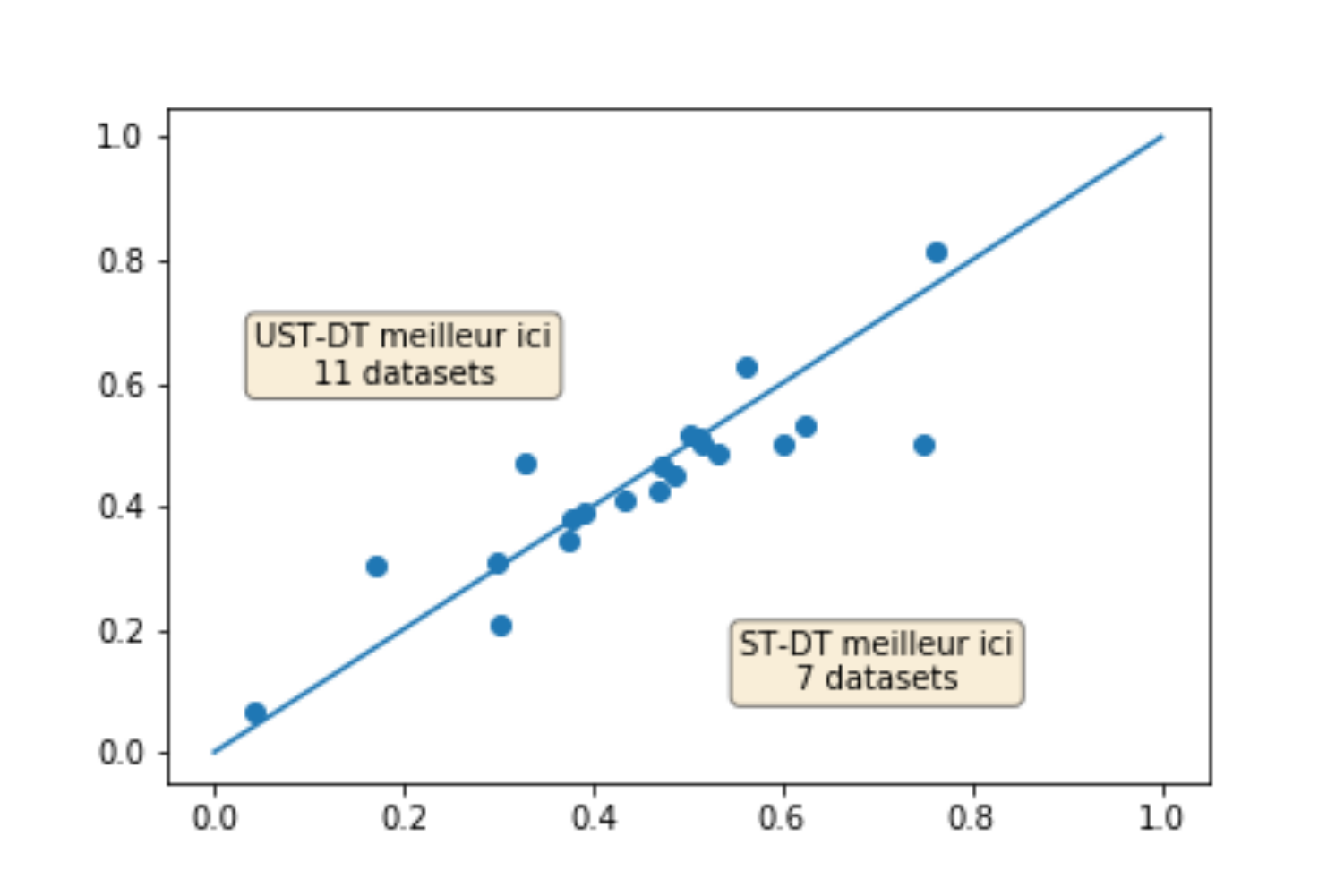}
		\caption{UST-DT vs ST-DT}
		\label{fig:scatter_plot_ust_vs_st_dt_gauss}
	\end{subfigure}
	\caption{Taux de classification en utilisant des lois normales}
\end{figure}

Dans la seconde façon de gérer l'incertitude, nous utilisons des lois normales; chaque composante $j$ d'un vecteur caractéristique est remplacée par la probabilité de réalisation de l'estimation optimiste. Cette probabilité est donnée par une loi gaussienne de moyenne $\frac{max_j + min_j}{2}$ et d'écart-type $\frac{1}{\sqrt{2\pi}}$. $min_j$ et $max_j$ sont respectivement les distances minimales et maximales  au shapelet $j$ sur le jeux d'apprentissage. Comme précédemment, UST est meilleur avec les forêts de rotation qu'avec les arbres de décision (figure \ref{fig:scatter_plot_ust_dt_vs_rotf_gauss}). Cependant, cette façon de prendre en compte l'incertitude rend UST meilleur que ST. En effet UST-DT l'emporte sur $11$ jeux de données et perd sur $7$. 

\begin{figure}
	\centering
	\begin{subfigure}{.5\textwidth}
		\centering
		\includegraphics[width=\linewidth]{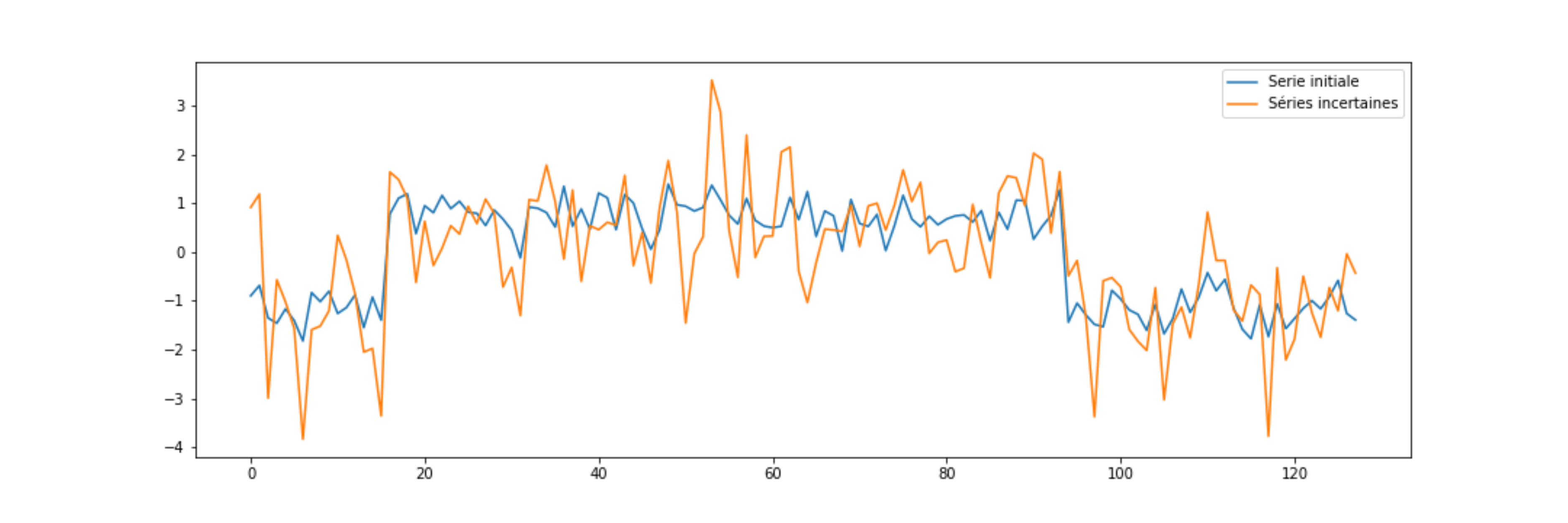}
		\caption{Jeu de données CBF}
	\end{subfigure}%
	\begin{subfigure}{.5\textwidth}
		\centering
		\includegraphics[width=\linewidth]{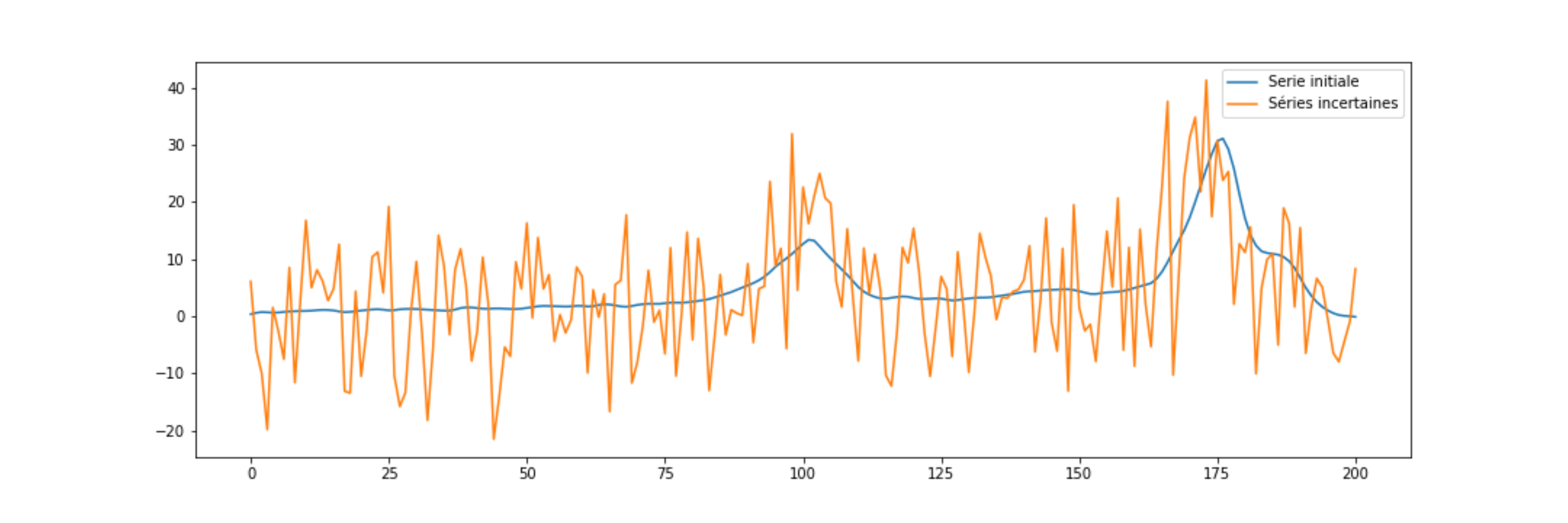}
		\caption{Jeu de données Fungi}
	\end{subfigure}
	\caption{Illustration de l'incertitude sur deux jeux de données}
	\label{fig:uncertainty_illustration}
\end{figure}

En observant de plus prêt les jeux de données sur lesquels nos modèles sont très mauvais, on se rend compte que l'incertitude est très élevée dans ces jeux de données. La figure \ref{fig:uncertainty_illustration} illustre l'effet de l'incertitude sur deux instances provenant des jeux de données Fungi et CBF. Ce sont là les deux jeux de données sur lesquels nos modèles sont très mauvais. Cependant, ceci n'est pas très surprenant car les modèles de classifications de séries temporelles sont très vulnérables aux attaques \citep{fawaz2019adversarial,karim2019adversarial}.

\section{Conclusion}\label{sec:conclusion}
Dans cet article, nous avons présenté une adaptation de la transformation en shapelet pour la classification de séries temporelles incertaines. Pour cela nous avons utilisé \textit{UDISSIM}, une nouvelle mesure de dissimilarité pour des vecteurs vecteurs incertains. Les expérimentations ont montré que la propagation de l'incertitude dans la transformation shapelet permet d'avoir un meilleur taux de classification; Mais pour cela il faut que l'incertitude propagée soit bien gérée dans la phase de classification. En perspective nous comptons utiliser un modèle de classification supervisé qui soit réellement conscient de l'incertitude. En effet nous pensons qu'on aurait de meilleurs résultats en prenant compte de l'incertitude durant la phase d'entraînement du modèle de classification. Afin d'évaluer davantage notre approche, nous comptons ensuite faire la transformation shapelet en utilisation respectivement PROUD, DUST et MUNICH. 
\bibliographystyle{rnti.bst}
\bibliography{egc2020}

\Fr

\end{document}